\documentclass[runningheads]{llncs}
\usepackage[T1]{fontenc}
\usepackage{graphicx}
\usepackage{tablefootnote}
\usepackage{siunitx}
\usepackage[colorlinks=true, allcolors=blue]{hyperref}
\usepackage{color}

\begin{document}
\title{Advancing STT for Low-Resource Real-World Speech}
%
\author{Flavio D'Intino\inst{1}\orcidID{0009-0007-2245-7799} \and
Hans-Peter Hutter\inst{1}\orcidID{0000-0002-1709-546X}}
\authorrunning{F. D'Intino \and H.-P. Hutter}
\institute{Institute of Computer Science, Zurich University of Applied Sciences, Winterthur, Switzerland \\
\email{\{dint,huhp\}@zhaw.ch}}
\maketitle
\begin{abstract}
Swiss German is a low-resource language represented by diverse dialects that differ significantly from Standard German and from each other, lacking a standardized written form. As a result, transcribing Swiss German involves translating into Standard German. Existing datasets have been collected in controlled environments, yielding effective speech-to-text (STT) models, but these models struggle with spontaneous conversational speech.

This paper, therefore, introduces the new SRB-300 dataset, a 300-hour annotated speech corpus featuring real-world long-audio recordings from 39 Swiss German radio and TV stations. It captures spontaneous speech across all major Swiss dialects recorded in various realistic environments and overcomes the limitation of prior sentence-level corpora.

We fine-tuned multiple OpenAI Whisper models on the SRB-300 dataset, achieving notable enhancements over previous zero-shot performance metrics. Improvements in word error rate (WER) ranged from 19\% to 33\%, while BLEU scores increased between 8\% and 40\%. The best fine-tuned model, large-v3, achieved a WER of 17.1\% and a BLEU score of 74.8. This advancement is crucial for developing effective and robust STT systems for Swiss German and other low-resource languages in real-world contexts.

\end{abstract}
\section{Introduction}
\label{sec:intro}
Swiss German, a collection of Alemannic dialects spoken by about 3.9 million people in Switzerland,\footnote{\href{https://www.bfs.admin.ch/bfs/de/home/statistiken/kataloge-datenbanken.assetdetail.31085875.html}{Bundesamt für Statistik: Hauptsprachen seit 1910}, accessed 29.07.2024} is known for its diversity with hundreds of local dialects varying significantly across regions.\footnote{\href{https://www.eda.admin.ch/aboutswitzerland/en/home/gesellschaft/sprachen/sprachen-und-dialekte.html}{Federal Departement of Foreign Affairs - Languages and dialects}, accessed 09.10.2024} Swiss German significantly differs from Standard German in phonetics, vocabulary, morphology, and syntax, making it difficult for German listeners to understand without training.

Unlike Standard German, Swiss German lacks a standardized written form and is typically only used in informal written contexts, such as text messages. Therefore, transcribing Swiss German is conventionally done in Standard German, which is considered a translation task rather than a mere transcription. This translation comes with the corresponding difficulties, e.g., due to spelling ambiguities or only partly matching vocabularies~\cite{paonessa_dialect_2023}. Phonetic, morphological, and syntactic variations across Swiss German dialects and even within the same dialect further complicate the transcription task.

Section~\ref{sec:related} provides an overview of the most common speech corpora for Swiss German and the SOTA STT models trained with these datasets. In Section~\ref{sec:datasets}, we discuss the details of these datasets and highlight their limitations, as well as the limitations of the SOTA models in real-world transcription tasks. Section~\ref{sec:new-corpus} presents the specifics of the new SRB-300 speech corpus and the methodology used for its collection. In Section~\ref{sec:experiments}, we explain how we fine-tuned various OpenAI Whisper models using SRB-300 for practical applications in low-resource languages, specifically Swiss German. Finally, Section~\ref{sec:results} compares the performance of the different fine-tuned Whisper models against other SOTA models on the new SRB-300 corpus.

\section{Related Work}
\label{sec:related}
Several efforts have been made to collect Swiss German speech corpora for training and optimizing STT models~\cite{dogan-schonberger_swissdial_2021,pluss_stt4sg-350_2023,pluss_sds-200_2022,pluss_swiss_2021}. These datasets were gathered in relatively controlled settings, such as partially read speech, indoor environments, single speakers, and minimal background noise. Using these datasets, various deep learning-based STT models for Swiss German have been trained, including Conformer~\cite{gulati_conformer_2020} and the foundation model XLS-R~\cite{babu_xls-r_2022}. These models have performed well on their respective test sets~\cite{pluss_stt4sg-350_2023,pluss_swiss_2021}, as summarized in Section~\ref{sec:datasets}.

Similarly, OpenAI's Whisper models~\cite{radford_robust_2023} have also been evaluated for Swiss German~\cite{dolev_does_2024,sicard_spaiche_2023} using these datasets. The Whisper models showed surprisingly good zero-shot performance, although they were slightly less effective than models specifically trained on the Swiss German datasets. Recently, Whisper models have been fine-tuned on these datasets, as described in Section~\ref{sec:datasets}, employing different sample concatenation strategies~\cite{timmel_fine-tuning_2024}. While the fine-tuned models outperformed the previously best models on the available datasets, they still face challenges with more realistic speech data.

\section{Existing Datasets for Swiss German}
\label{sec:datasets}
Several publicly available speech corpora for Swiss German have been collected in recent years.

\subsubsection{SwissDial}
The SwissDial parallel corpus~\cite{dogan-schonberger_swissdial_2021} consists of 26 hours of single-sentence samples. Each sentence is spoken by 8 different speakers, each from a different region of Switzerland, resulting in approximately 3 hours of recordings for each dialect. The origin of the utterances is written texts in Standard German, including news articles, weather reports, and Wikipedia entries. The duration of the samples varies from 0.5\,s to 13.5\,s, with an average duration of 3.3\,s.

\subsubsection{SPC}
The Swiss Parliaments Corpus (SPC)~\cite{pluss_swiss_2021} consists of 299 hours of transcribed recordings of the Bernese parliament "Grosser Rat". The data is divided into a training set of 293 hours and a test set of 6 hours. Since the parliament is located in Bern, most speakers use a Bernese dialect. The annotations are not entirely accurate transcriptions. The duration of the samples varies from 0.1\,s to 108\,s, with an average duration of 7.1\,s.

\subsubsection{SDS-200}
The SDS-200 dataset~\cite{pluss_sds-200_2022} comprises 200 hours of spoken audio recorded by approximately 4000 speakers. However, the number of recordings varies significantly among speakers, following a long-tail distribution. The test set contains 5 hours of audio.
An online service was utilized in which speakers were presented with a Standard German sentence to be spoken in their respective Swiss German dialects. To ensure quality, the recordings were validated by other participants. The sentences were selected from Swiss newspapers and the Standard German dataset of the Common Voice corpus.\footnote{\url{https://commonvoice.mozilla.org}} The sample durations range from 2.0\,s to 11.2\,s, with an average duration of 4.8\,s.

\subsubsection{STT4SG-350}
The STT4SG-350~\cite{pluss_stt4sg-350_2023} consists of 343 hours of speech recorded by 316 different speakers, with the test set containing 34 hours. This dataset was collected in the same manner as the SDS-200 dataset. The written sentences are derived from news articles and recordings of the Swiss parliament. The sample durations range from 2.0\,s to 15.4\,s, with an average duration of 5.0\,s.

\subsection{Limitations of these Datasets}
\label{subsec:limitations}
The aforementioned corpora are particularly valuable for researching STT models and benchmarking purposes. However, models trained on these datasets face challenges when transcribing realistic, spontaneously spoken Swiss German, particularly in everyday conversational situations. These datasets have one or more of the following limitations:
\begin{itemize}
    \item short samples (a few seconds only) with a single utterance
    \item controlled recordings in quiet environments
    \item few spontaneous speech: utterances either translated from Standard German sentences or read from a manuscript, therefore hardly any spontaneous speech characteristics (disfluencies, hesitations, slip of the tongue, repetitions, etc.)
    \item only one single speaker per sample
\end{itemize}

The SOTA STT models trained with these datasets perform exceptionally well on the corresponding test sets (see Table~\ref{tab:ref_performance}). However, their performance dramatically drops on more realistic, out-of-distribution datasets, such as the new SRB-300 corpus (see Table~\ref{tab:srb-300_results}).
\begin{table}
    \caption{Reported results of the fine-tuned SOTA models on the test sets of the most common scientific corpora for Swiss German. For reference, we include the zero-shot results of the pre-trained XLS-R 1B model on the new SRB-300 corpus, which is introduced in this paper.}
    \label{tab:ref_performance}
    \centering
    \begin{tabular}{|l|c|c|l|l|}
        \hline
        Corpus & WER & BLEU & fine-tuned Model \\
        \hline
         SPC        & 23.7 & 60.7 & XLS-R 1B, \cite{schraner_swiss_2022} \\
         SDS-200    & 18.2 & 69.6 & XLS-R 1B, \cite{pluss_stt4sg-350_2023} \\
         STT4SG-350 & 14.0 & 74.7 & XLS-R 1B, \cite{pluss_stt4sg-350_2023} \\
         SRB-300    & 44.4 & 37.5 & XLS-R 1B, this work (ref. Table~\ref{tab:srb-300_results}) \\
         \hline
    \end{tabular}
\end{table}
The short samples in these datasets pose significant challenges for training models like Whisper, which require an input length of 30 seconds. This necessitates substantial zero-padding for each training sample, leading to difficulties when the trained model is applied to longer audio sequences. Timmel et al. recently tackled this issue by combining several independent samples from the available datasets into 30-second samples using various strategies. Their resulting model surpassed the previous SOTA models across all datasets, yet it remained inferior to the original Whisper model when tested on more realistic, longer audio data~\cite{timmel_fine-tuning_2024}.

These findings highlight the necessity of addressing the aforementioned dataset limitations to enhance the performance of models on conversational Swiss German speech in various real-world contexts. To this end, we have compiled a new corpus specifically addressing these limitations.

\section{New Real-World Speech Corpus for Swiss German}
\label{sec:new-corpus}
We introduce the new SRB-300 (300\,h Swiss Regional Broadcasts) dataset, an annotated speech corpus that addresses the limitations of previous Swiss German datasets. The SRB-300 dataset includes 303 hours of audio recordings from 39 regional Swiss German radio and TV broadcast stations across German-speaking Switzerland. It captures realistic and spontaneous speech in a variety of contexts, such as news reports, interviews, discussions, and live events from both indoor and outdoor settings. This dataset offers a representative selection of all major Swiss dialects and is designed for training with long audio samples. The following sections detail the dataset and describe the collection process.

\subsection{Data Collection and Pre-Processing}
The speech data was recorded from radio and TV broadcasts. To extract 30\,s speech samples for the datasets, the following steps were performed.
\begin{enumerate}
    \item Extraction: extract speech segments from the original audio
    \item Pre-transcription: automatically pre-transcribe the extracted speech audio
    \item Correction: manually correct the pre-transcription according to given rules
    \item Sample generation: chunk utterances into up to 30\,s long samples
\end{enumerate}

\subsubsection{Extraction}
Each available MP4 recording from 29 radio and 10 TV stations contained approximately 8-hour excerpts of a day's radio and TV broadcast, including non-speech sections, resulting in a total of about 1790 hours of audio. In the first step, irrelevant parts, such as music, jingles, commercials, and repeated programs, were identified and removed from the recordings. Professionals then manually labeled the remaining speech data with contextual metadata, including details like speaker gender and type of broadcast. One-third of the data comes from TV broadcasts, while two-thirds comes from radio (ref. Appendix~\ref{app:datasets}, Table~\ref{tab:data_per_media_type}).

\subsubsection{Pre-Transcription}
The approximately 300 hours of speech from the extraction step were automatically pre-transcribed using OpenAI's Whisper large-v3\footnote{\url{https://huggingface.co/openai/whisper-large-v3}} model. The pre-transcribed utterances were then joined into 28-second speech audio chunks using the following process:
\begin{itemize}
    \item Successive speech segments were merged into whole utterances (usually a sentence) by merging their timestamps and transcripts. For example, the two pre-transcribed segments:
\begin{verbatim}
(307.62, 300.7)	Und auch am Mittag und Nachmittag ist 
                der Himmel meistens blau,
(300.7, 312.3)	häufig sogar wolkenlos über dem Jura.
\end{verbatim}
were merged into the following utterance
\begin{verbatim}
(307.62, 312.3)	Und auch am Mittag und Nachmittag ist
                der Himmel meistens blau, häufig sogar 
                wolkenlos über dem Jura.
\end{verbatim}
    \item Timestamps were rounded to whole seconds because the transcription tool used could only handle integer timestamps.
    \item The German character `ß` was replaced with 'ss', as it does not exist in Swiss German.
    \item Successive utterances were merged to segments of up to 28\,s.\footnote{Finally, we aimed for a maximum sample length of 30\,s. Therefore, we left some margins so that the segment boundaries could be adjusted during the later manual correction step.} 
\end{itemize}
The last step involved creating XML files that included the combined transcriptions. These files, along with the audio files, served as the input data for the transcription tool used in the next step. 

\subsubsection{Manual Correction}
The pre-transcriptions from the previous phase were manually reviewed and adjusted according to predefined rules (see Appendix~\ref{app:transcription_rules}). Along with the transcription text, the start and end timestamps of the segments were revised to ensure that the entire audio content of the transcription was included as accurately as possible.

By providing automated pre-transcriptions, the transcription time required for lay transcribers was reduced by a factor of 2 to 3. On average, the time to check and correct the transcription was 3 to 4 times the audio duration.

\subsubsection{Sample Generation}
The samples were extended by 0.2 seconds at both the beginning and the end to ensure that no utterances were inadvertently cut off due to inaccuracies in the rounded timestamps. Each sample was assigned a random identifier and automatically supplemented with metadata, including details such as the broadcast station, speaker gender, and its position (index) within the entire broadcast. A detailed description of each available attribute is provided in Table~\ref{tab:columns}. This data was obtained from the original metadata accompanying the raw audio data.

\begin{figure}
    \includegraphics[width=\textwidth]{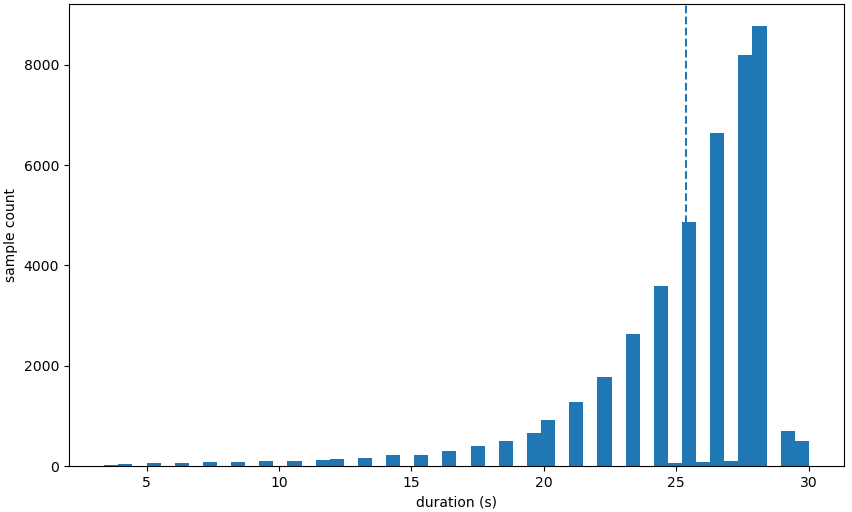}
    \caption{Distribution of sample duration over all datasets in SRB-300. The durations range from 3.4\,s to 30.0\,s, with an average of 25.4\,s. Each of the three datasets (ref. Section~\ref{sec:partition}) has a similar distribution.}
    \label{fig:sample-length-distribution}
\end{figure}

The sample generation process resulted in the sample length distribution depicted in Figure~\ref{fig:sample-length-distribution}. Thus, the samples are considerably longer than in the corpora in section~\ref{sec:datasets}. It is important to note that samples with lengths significantly shorter than 30 seconds typically occur at the end of the original audio file or cannot be merged with adjacent segments, as doing so would lead to sample lengths exceeding 30 seconds.
\begin{table}
   \centering
    \caption{Metadata attributes of the samples in the SRB-300 dataset.}
    \label{tab:columns}
    \begin{tabular}{|l|l|}
        \hline
        Metadata & Description \\
        \hline
        clip\_id & unique clip identifier \\
        clip\_path & path to the Swiss German clip in the audio folder \\
        sender\_id & unique identification of the broadcast station \\
        sample\_id & unique sample identifier \\
        duration\_s & duration of the clip in seconds \\
        starts & start timestamps of the merged utterances in the original audio \\
        ends & end timestamps of the merged utterances in the original audio \\
        original\_media\_file & original audio file, from which the sample was extracted \\
        index & position of the given sample in the original media file \\
        sender & name of the broadcast station \\
        recording\_date & date on which the sample was recorded from broadcasting \\
        gender & gender(s) of the speaker(s) \\
        speaker & kind of speaker (presenter, news reader, reporter, ...) \\
        form & live, story, report, talk \\
        type & information, moderation, service, entertainment, ... \\
        kind & information, moderation, traffic, weather, ... \\
        style\_element & statement, interview, still image, ... \\
        language & language spoken in the sample \\
        media\_type &type of broadcasting (usually radio, TV, or social media) \\
        text & Standard German text (transcript) \\
        \hline
    \end{tabular}
\end{table}

\begin{figure}
    \includegraphics[width=\textwidth]{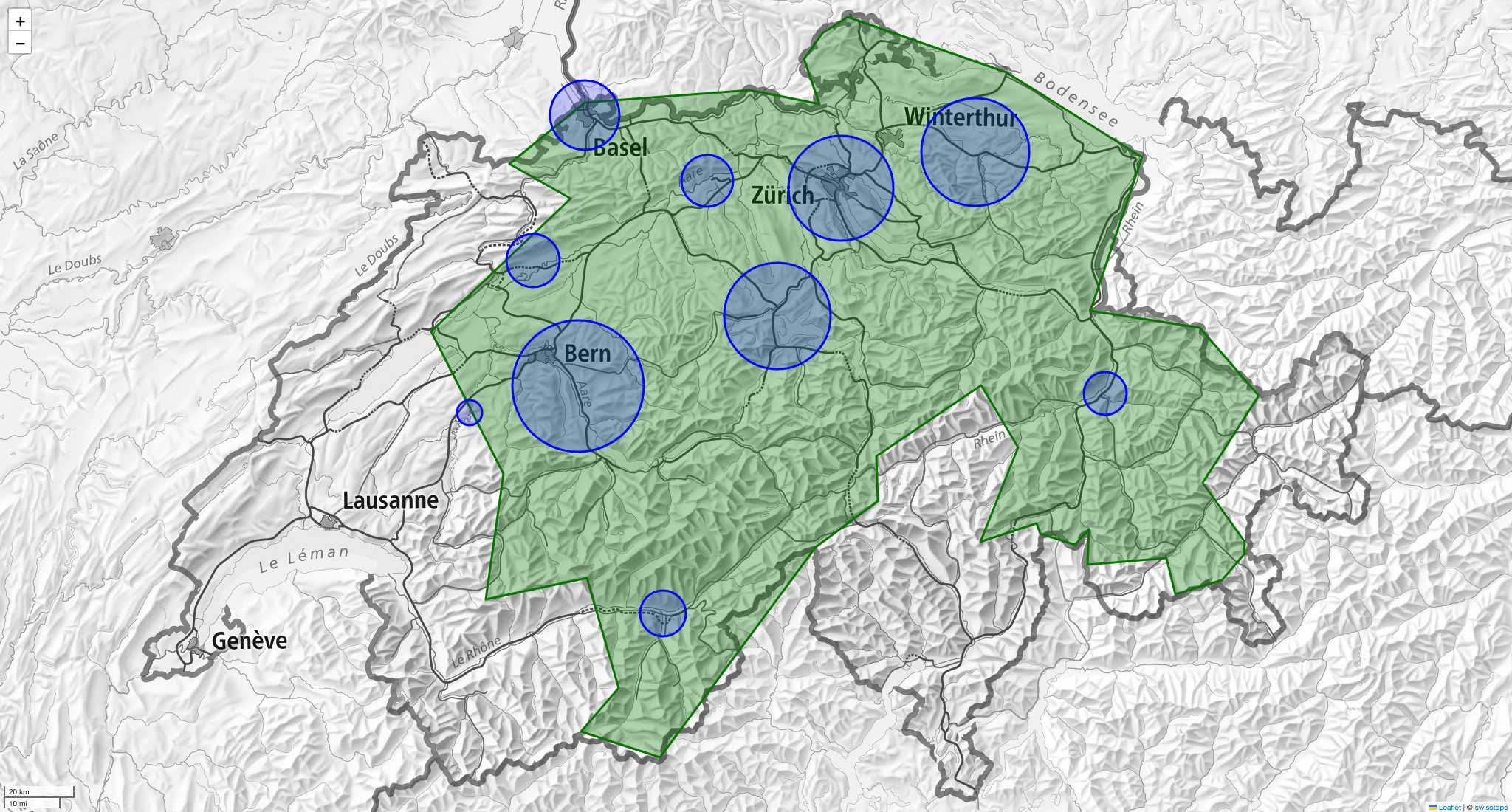}
    \caption{Distribution of samples over Switzerlands German-speaking area}
    \label{fig:data_per_region}
\end{figure}

\begin{figure}
    \includegraphics[width=\textwidth]{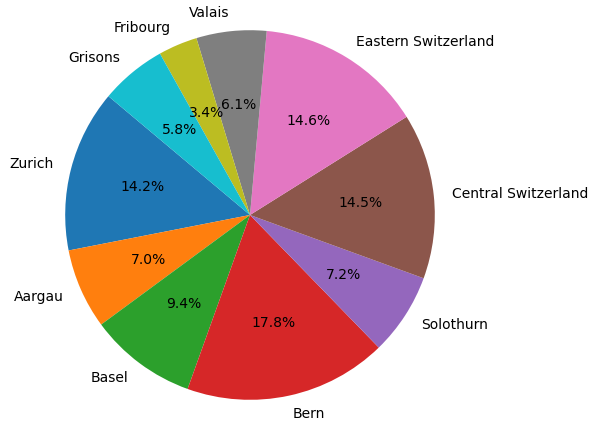}
    \caption{Distribution of the total sample duration per dialect region}
    \label{fig:data_per_region_pi}
\end{figure}

Figure~\ref{fig:data_per_region} shows the geographic distribution of the broadcasting stations, and Figure~\ref{fig:data_per_region_pi} the length distribution of the audio samples in the SRB-300 dataset. To that end, the broadcasting stations were categorized into 10 dialect regions. The diameters of the circles in Figure~\ref{fig:data_per_region} represent the total duration of samples collected from broadcast stations within each region. 
We have ensured that the SRB-300 corpus is weighted to reflect the size of their speaker populations.\footnote{The majority of the Swiss population lives in the "Mittelland," located north of the Alps between Geneva and Lake Constance.} Since regional broadcasting stations generally feature local presenters and cater to local listeners, we can assume that the dialects of the speakers closely align with the locations of these stations. Therefore, we anticipate that the most prominent Swiss dialects are well represented in the dataset.
The green background in Figure~\ref{fig:data_per_region} indicates the approximate border of the Swiss German language in Switzerland.\footnote{The Swiss German language border is approximated according to \href{https://www.atlas.bfs.admin.ch/maps/13/de/17138_17137_235_227/26599.html}{Die 4 Sprachgebiete der Schweiz nach Gemeinden, 2020}, accessed 18.09.2024}

The dataset was not artificially gender-balanced, but based on the available metadata, we estimate a male-to-female speaker ratio of approximately 2:1 (ref. Table~\ref{tab:data_per_speaker_gender} in Appendix~\ref{app:datasets}).

The broadcasts cover a wide variety of formats, including entertainment, information, news, traffic updates, weather reports, and moderation in both indoor and outdoor settings (see Appendix~\ref{app:datasets}, Tables~\ref{tab:data_per_post_kind} and~\ref{tab:data_per_journalistic_form}). Additional statistical details about the SRB-300 corpus are in Appendix~\ref{app:datasets}.

\subsection{Dataset Partitioning}
\label{sec:partition}
The SRB-300 speech corpus is divided into three datasets: training (76\%), validation (10\%), and test (14\%), as shown in Table~\ref{tab:sample_count_incl_out_of_training}. The test set also includes 6 hours of audio from 6 radio stations not present in the training or validation datasets. This allows us to evaluate performance on previously unseen broadcast stations. 

The samples in all datasets are organized according to their original order in the broadcasts, which allows for long-audio tests by concatenating sequential samples from the same broadcast. Each sample appears in only one dataset, but some broadcasts may include samples in different datasets. Additionally, samples from the same speaker may be present in multiple datasets. This overlap happens because our metadata lacks unique speaker labels.
\begin{table}
   \centering
    \caption{Sample counts and audio duration of the SRB-300 datasets}
    \label{tab:sample_count_incl_out_of_training}
    \begin{tabular}{|l|c|c|c|c|}
        \hline
        Dataset & Num. samples& Ratio (\%)& t (h) & t rel. (\%) \\
        \hline
        Training & 32714 & 76 & 230 & 76\\
        Validation & 4078 & 9 & 29 & 10\\
        Test & 6315 & 15& 44 & 14 \\\hline
         & \textbf{43107} &  & \textbf{303} & \\
        \hline
    \end{tabular}
\end{table}

\section{Experiments}
\label{sec:experiments}
We used different sizes of OpenAI's Whisper\footnote{\url{https://github.com/openai/whisper}} model for our fine-tuning experiments. Whisper is an open-source multilingual transformer~\cite{vaswani_attention_2017} model designed for automatic speech recognition, translation, and language detection~\cite{radford_robust_2023}. It was trained on an extensive multilingual corpus collected through web crawling. Thanks to its internal segmentation, Whisper can transcribe audio of arbitrary length~\cite{radford_robust_2023}.

Although Swiss German is not officially included in Whisper’s training data, we, along with other researchers, have identified indications of Swiss broadcasts in the training data~\cite{timmel_fine-tuning_2024}. This observation is supported by some hallucinated outputs generated by the Whisper models. This may help explain why Whisper models performed surprisingly well in several zero-shot experiments involving Swiss German transcription~\cite{dolev_does_2024,sicard_spaiche_2023,timmel_fine-tuning_2024}.

There are several established approaches for fine-tuning Whisper models \cite{deschamps-berger_asr-whisper-finetuning_2024,gandhi_fine-tune_2022,lodagala_fine-tuning_2023,ma_adapting_2023,srivastav_faster_2023}. Some specifically focus on low-resource languages~\cite{do_using_2023,ferraz_multilingual_2024,hsu_meta-whisper_2024,liu_exploration_2024,pillai_multistage_2024,pineiro-martin_weighted_2024,qian_learn_2024}.
Our fine-tuning process was based on the method described by Sanchit Gandhi~\cite{gandhi_fine-tune_2022} utilizing the Hugging Face implementation of Whisper as part of their transformers-library~\cite{wolf_transformers_2020} at version 4.46 with PyTorch 2.4, along with several Hugging Face libraries to facilitate multi-GPU training. This approach has proven effective for low-resource languages~\cite{liu_exploration_2024}.

We trained the Whisper models using 2 NVIDIA A100 40 GB GPUs for 4 to 12 epochs. We set a batch size of 4, expanded through gradient accumulation over 32 steps. 
We utilized an AdamW-optimizer~\cite{loshchilov_decoupled_2019}.
Furthermore, we applied a weight decay of 1\%.
The initial learning rate was set to \(5\times e^{-6}\) for small and medium model sizes. However, for the large model sizes, we reduced the initial learning rate to \(1\times e^{-6}\). To balance the training speed and memory requirements, we employed gradient checkpointing~\cite{chen_training_2016}.

Since Swiss German is unavailable as a configurable language in Whisper, we used the German language tag (DE) instead.

As a reference point, we compared our results against a SOTA XLS-R 1B model, which was trained on the STT4SG-350 dataset~\cite{pluss_stt4sg-350_2023}. Our experiments utilized the Hugging Face implementation for XLS-R.

\section{Results}
\label{sec:results}
In our comparisons, we calculated WER and BLEU score~\cite{papineni_bleu_2002}, commonly used for machine translation tasks, which we consider more relevant for Swiss German to Standard German transcription (ref. Section~\ref{sec:intro}). For the WER computation, we utilized the evaluate-library\footnote{\url{https://github.com/huggingface/evaluate}} of Hugging Face in version 0.4, which internally uses the jiwer\footnote{\url{https://github.com/jitsi/jiwer}} implementation.
For the BLEU score calculation, we used the NLTK~\cite{bird_steven_natural_2009} implementation, version 3.9. Before calculating any metrics, we applied Whisper's basic text normalizer to both the output text and the ground truth to ensure uniformity. The normalizer converts the text to lowercase, for example.
\begin{table}
    \caption{Results of different models on the SRB-300 test set using the Hugging Face implementation of Whisper with a beam width of 2.}
    \centering
    \label{tab:srb-300_results}
    \begin{tabular}{|l|l|c|c|}
        \hline
         Model & Fine-Tuning & WER & BLEU \\
         \hline
         Whisper small   & zero-shot & 37.4 & 45.3 \\ 
         ZHAW small   &   on SRB-300 & 25.0 & 63.2 \\
         \hline
         Whisper medium  & zero-shot & 27.2 & 58.1 \\
         ZHAW medium  &   on SRB-300 & 19.7 & 70.5 \\ 
         ZHAW medium 2  & on STT4SG-350 & 31.4 & 50.5 \\
         ZHAW medium 3 & on STT4SG-350 and SRB-300 & 19.4 & 71.0 \\
         \hline
         Whisper large-v2 & zero-shot & 23.5 & 63.7 \\
         ZHAW large-v2 & on SRB-300   & 18.4 & 72.3 \\
         \hline
         Whisper large-v3 & zero-shot & 21.1 & 69.1 \\
         ZHAW large-v3 & on SRB-300   & \textbf{17.1} & \textbf{74.8} \\
        \hline
         Whisper large-v3-turbo & zero-shot & 26.5 & 61.8 \\
         ZHAW large-v3-turbo & on SRB-300    & 18.7 & 71.8 \\
        \hline
        XLS-R-1B & on STT4SG-350 & 44.4 & 37.5 \\
        XLS-R-1B & on STT4SG-350 and SRB-300  & 24.7 & 60.8 \\
        \hline
    \end{tabular}
\end{table}
Table~\ref{tab:srb-300_results} summarizes the key findings from our experiments. It indicates that fine-tuning with the SRB-300 training set led to a reduction in the Whisper zero-shot WER for all models, with improvements ranging from 19\% for Whisper large-v3 to 33\% for Whisper small. Additionally, the BLEU scores increased by 8\% for Whisper large-v3 and up to 40\% for Whisper small. The best performance was achieved with the fine-tuned Whisper large-v3 model (referred to as ZHAW large-v3), which achieved a WER of 17.1\% and a BLEU score of 74.8 on the SRB-300 test set.

Manual inspection of the transcriptions indicated that they were generally highly accurate. Most errors were attributed to uncommon named entities, such as locations, brands, and persons.

\begin{figure}
    \includegraphics[width=\textwidth]{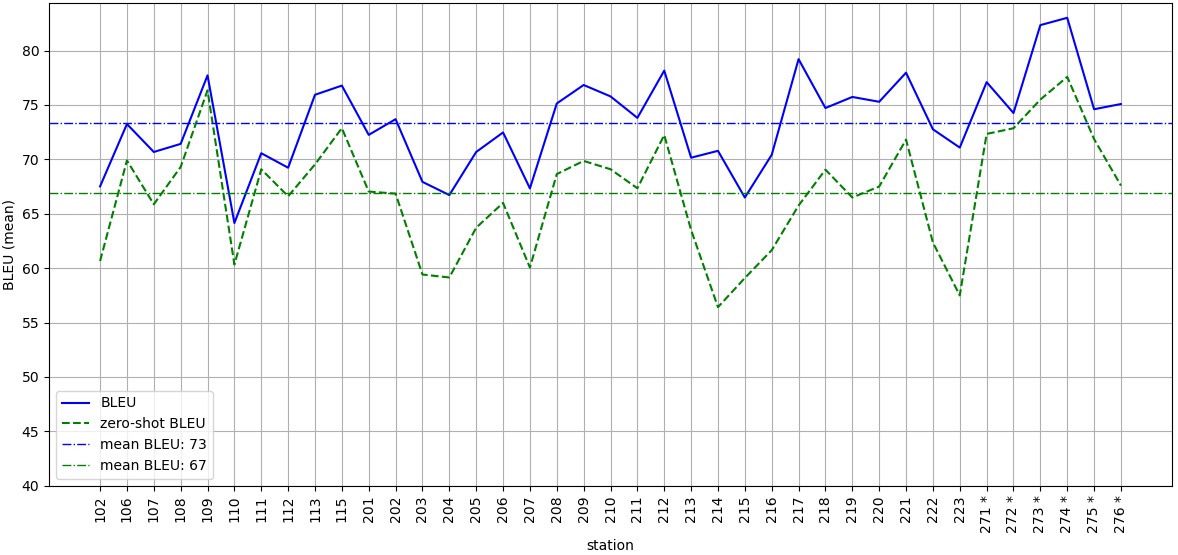}
    \caption{Improvement of mean BLEU score on the SRB-300 test set for all samples of each broadcasting station for the ZHAW large-v3 model, which was fine-tuned on SRB-300. The broadcast stations not in the training set are marked with a star. The y-axis begins at 40 for better readability.}
    \label{fig:bleu_per_station_large}
\end{figure}

Further analysis of the results (see Figure~\ref{fig:bleu_per_station_large}) revealed that all dialects in the training data benefited from the fine-tuning, despite a significant variance in the quantity of training data available for the different broadcasting stations, which ranged from less than 2 hours to over 11 hours (ref. Table~\ref{tab:data_per_broadcast_station}). Also, the broadcasting stations with no training data improved significantly (marked with a star in Figure~\ref{fig:bleu_per_station_large}).

Meta's\footnote{Formerly Facebook} XLS-R 1B model~\cite{babu_xls-r_2022}, which was specifically fine-tuned on Swiss German using the STT4SG-350 dataset, demonstrated a significant performance drop from the reported 14.0\% WER and 74.7 BLEU~\cite{pluss_stt4sg-350_2023} down to 44.4\% WER and 37.5 BLEU score on the SRB-300 test set. This represents a 50\% worse performance in BLEU score compared to the best fine-tuned Whisper model ZHAW large-v3. Surprisingly, its performance was considerably lower than the zero-shot performance of Whisper's medium model, which achieved a WER of 27.2\% and a BLEU score of 58.1 on the SRB-300 test set. This underscores the necessity of utilizing realistic datasets for STT model training.

Additionally, we fine-tuned the XLS-R 1B model on the new SRB-300 dataset. This resulted in performance improvements comparable to the fine-tuned ZHAW small model. 

In two additional experiments, we aimed to determine how Whisper models might benefit from fine-tuning using the sentence-level STT4SG-350 dataset for our specific task. We fine-tuned the medium-sized Whisper model on the STT4SG-350 dataset and evaluated its performance on the SRB-300 test set (experiment ZHAW medium 2). However, we observed a 4.2\% increase in the WER and a drop of 7.6 BLEU score compared to the Whisper medium model's zero-shot performance. After conducting further fine-tuning with the SRB-300 training set (referred to as ZHAW medium 3), the model’s performance only slightly improved compared to the Whisper model that had been fine-tuned solely on the SRB-300 dataset. This suggests that additional fine-tuning data only helps if recorded in a setting similar to the target application.

The final experiment evaluated the latest Whisper model, large-v3-turbo,\footnote{\url{https://github.com/openai/whisper/discussions/2363}, accessed 19.11.2024} using the SRB-300 dataset. After fine-tuning with SRB-300, the model performed similarly to the large-v2 model, but its results were lower than those of the large-v3 model (see Table~\ref{tab:srb-300_results}). One of the key advantages of the large-v3-turbo model is its inference speed. It operates even faster than the small model size measured on the SRB-300 test set (see Table~\ref{tab:srb-300_times}). It is important to note that the reported inference times do not account for any preprocessing steps. The experiments were conducted using two NVIDIA A100 40 GB GPUs, with a batch size of 64 and a beam width of 2.

In a separate experiment, we concatenated consecutive test set samples to create complete audio broadcasts. The resulting test audio files reached lengths of up to 1 hour, with an average duration of 21 minutes. Using the ZHAW medium and ZHAW large-v3 models, we found that the performance is similar to the results presented in Table~\ref{tab:srb-300_results}. This indicates that the findings in Table~\ref{tab:srb-300_results} remain valid even for audio segments considerably longer than those typically found in scientific datasets.
\begin{table}
\centering
    \caption{Inference times on the SRB-300 test set}
    \label{tab:srb-300_times}
    \begin{tabular}{|l|c|c|}
    \hline
    Model & Inference time (min.)& Ratio to large-v3 (\%) \\
    \hline
    Whisper small    & 29 & 63 \\
    Whisper medium   & 44 & 95 \\
    Whisper large-v2 & 47 & 101 \\
    Whisper large-v3 & 46 & 100 \\
    Whisper large-v3-turbo & 19 & 40 \\
    \hline
    \end{tabular}
\end{table}
\section{Conclusion}
The primary contributions of this paper are twofold. First, it introduces a new realistic speech corpus for Swiss German STT. Unlike many existing datasets for Swiss German, this new corpus includes long samples of spontaneous speech recorded in real-world scenarios, such as conversations with various background noises.

Second, the paper presents a robust STT model for conversational Swiss German, named ZHAW large-v3. This model performs 50\% better on the realistic SRB-300 test set compared to the best models that have been fine-tuned on the STT4SG-350 dataset. With a BLEU score of nearly 75, it performs similarly on the challenging SRB-300 test set as the best-reported model trained and tested on the scientific STT4SG-350 dataset.

These findings underscore the potential for fine-tuning multilingual STT models for low-resource languages and emphasize the importance of leveraging realistic data to enhance these models for real-world applications.

\section{Future Work}
Although the performance of the ZHAW large-v3 STT model is very promising, the remaining fallacies often relate to uncommon named entities. These entities are crucial in professional applications, particularly in legal and medical fields. Therefore, our future research will focus on fine-tuning our models to improve their handling of these named entities.

\section*{Limitations}
The SRB-300 corpus cannot be made publicly available due to data license restrictions of the industrial partner. For requests related to the SRB-300 corpus, please contact our industrial partner.\footnote{Eurospider Information Technology AG, \url{https://eurospider.com/}}

Apart from gender, we do not have detailed demographic information about the speakers of SRB-300. We only know the location of the broadcasting station and not the origins of the speakers. Local broadcasting stations tend to promote themselves by employing presenters who speak the local dialect. Thus, we assume that the dialects used by the speakers roughly correspond to the region where the station is located. Further uncertainties are detailed in Appendices~\ref{app:pii} and \ref{app:risks}.

\begin{credits}
\subsubsection{\ackname}
We thank our industrial project partners for providing the audio and metadata for the new SRB-300 corpus. Additionally, we appreciate the assistance of our students in reviewing and correcting the transcriptions. We also thank all our colleagues at ETHZ, FHNW, and ZHAW for making their datasets SwissDial, SPC, SDS-200, and STT4SG-350 available. Innosuisse funded this study under project number 105.791 IP-ICT.

\subsubsection{\discintname}
The authors have no competing interests to declare that are relevant to the content of this article.
\end{credits}
\appendix
\section{Dataset Characteristics}
\label{app:datasets}
Table~\ref{tab:data_per_media_type} lists the distribution of the recorded broadcast types. In Table~\ref{tab:data_per_speaker_gender}, the distribution of the speakers' gender can be seen. Table~\ref{tab:data_per_speaker} lists the various speaker types. Table~\ref{tab:data_per_broadcast_station} lists the distribution of broadcasting stations (anonymized). Table~\ref{tab:data_per_post_kind} outlines the distribution of post kind, and Table~\ref{tab:data_per_journalistic_form} details the journalistic forms of the posts. Each table provides the distribution of all data, as well as that of the training and test sets, respectively.
\begin{table}
    \caption{Broadcast type}
    \label{tab:data_per_media_type}
    \centering
    \begin{tabular}{|l|S|S|S|S|S|S|}
        \hline
        Media type & {Total (h)} & {Total (\%)} & {Training (h)} & {Training(\%)} & {Test (h)} & {Test(\%)} \\
        \hline
        radio & 198& 65& 149& 65& 30& 68\\
        tv & 106& 35& 81& 35& 14& 32\\
        \hline
    \end{tabular}
\end{table}
\begin{table}
    \caption{Available speaker gender information}
    \label{tab:data_per_speaker_gender}
    \centering
    \begin{tabular}{|l|S|S|S|S|S|S|}
        \hline
        Speaker Gender & {Total (h)} & {Total (\%)} & {Training (h)} & {Training(\%)} & {Test (h)} & {Test(\%)} \\
        \hline
        female & 70 & 23 & 55 & 24 & 9 & 20 \\
        female \& male & 49 & 16 & 39 & 17 & 5 & 11 \\
        male & 136 & 45 & 104 & 45 & 19 & 42 \\
        unknown & 48 & 16 & 32 & 14 & 12 & 26 \\
        \hline
    \end{tabular}
\end{table}
\begin{table}
    \caption{Data on speaker types}
    \label{tab:data_per_speaker}
    \centering
    \begin{tabular}{|l|S|S|S|S|S|S|}
        \hline
        Speaker & {Total (h)} & {Total (\%)} & {Training (h)} & {Training(\%)} & {Test (h)} & {Test(\%)} \\
        \hline
        correspondent & 0.2 & 0.1 & 0.2 & 0.1 & 0.0 & 0.0 \\
        expert & 2.6 & 0.8 & 2.1 & 0.9 & 0.3 & 0.7 \\
        informant & 31.9 & 10.5 & 23.9 & 10.4 & 5.1 & 11.6 \\
        listener & 0.3 & 0.1 & 0.2 & 0.1 & 0.1 & 0.3 \\
        mixed & 80.6 & 26.5 & 58.8 & 25.5 & 14.0 & 31.8 \\
        moderator & 53.3 & 17.6 & 41.8 & 18.1 & 6.7 & 15.3 \\
        multiple moderators & 14.8 & 4.9 & 11.3 & 4.9 & 1.5 & 3.4 \\
        multiple news anchors & 0.6 & 0.2 & 0.5 & 0.2 & 0.1 & 0.2 \\
        news anchor & 38.6 & 12.7 & 29.8 & 12.9 & 5.4 & 12.2 \\
        other & 0.1 & 0.0 & 0.1 & 0.0 & 0.0 & 0.0 \\
        reporter/ journalist & 35.2 & 11.6 & 27.0 & 11.7 & 4.9 & 11.1 \\
        specialist journalist & 0.3 & 0.1 & 0.2 & 0.1 & 0.1 & 0.2 \\
        unknown & 45.0 & 14.8 & 34.9 & 15.1 & 5.8 & 13.1 \\
        \hline
    \end{tabular}
\end{table}
\begin{table}
    \caption{Audio data per broadcast station. IDs that start with 1 correspond to TV stations, while those starting with 2 correspond to radio stations. Stations 271 to 276 are excluded from the training set.}
    \label{tab:data_per_broadcast_station}
    \centering
    \begin{tabular}{|p{1.5cm}|S|S|S|S|S|S|}
        \hline
        Broadcast Station ID & {Total (h)} & {Total (\%)} & {Training (h)} & {Training(\%)} & {Test (h)} & {Test(\%)} \\
        \hline
        102 & 9.92 & 3.27 & 7.94 & 3.44 & 0.99 & 2.24 \\
        106 & 13.41 & 4.42 & 10.73 & 4.65 & 1.34 & 3.03 \\
        107 & 14.21 & 4.68 & 11.37 & 4.93 & 1.42 & 3.21 \\
        108 & 11.72 & 3.86 & 9.38 & 4.07 & 1.16 & 2.63 \\
        109 & 8.81 & 2.90 & 6.15 & 2.67 & 1.88 & 4.27 \\
        110 & 12.50 & 4.12 & 9.25 & 4.01 & 2.09 & 4.74 \\
        111 & 9.81 & 3.23 & 7.85 & 3.40 & 0.97 & 2.21 \\
        112 & 6.98 & 2.30 & 4.01 & 1.74 & 2.47 & 5.59 \\
        113 & 8.73 & 2.88 & 6.99 & 3.03 & 0.87 & 1.97 \\
        115 & 9.54 & 3.14 & 7.64 & 3.31 & 0.94 & 2.14 \\
        201 & 10.17 & 3.35 & 8.14 & 3.53 & 1.01 & 2.28 \\
        202 & 7.23 & 2.38 & 4.99 & 2.16 & 1.61 & 3.64 \\
        203 & 8.24 & 2.71 & 5.97 & 2.59 & 1.52 & 3.45 \\
        204 & 9.54 & 3.14 & 7.64 & 3.31 & 0.94 & 2.13 \\
        205 & 11.22 & 3.70 & 8.98 & 3.89 & 1.11 & 2.53 \\
        206 & 11.29 & 3.72 & 9.04 & 3.92 & 1.12 & 2.54 \\
        207 & 8.30 & 2.73 & 4.55 & 1.97 & 3.19 & 7.22 \\
        208 & 8.30 & 2.73 & 6.64 & 2.88 & 0.83 & 1.87 \\
        209 & 7.75 & 2.55 & 6.21 & 2.69 & 0.77 & 1.74 \\
        210 & 8.48 & 2.79 & 6.91 & 3.00 & 0.78 & 1.76 \\
        211 & 9.93 & 3.27 & 7.95 & 3.45 & 0.99 & 2.25 \\
        212 & 8.28 & 2.73 & 5.83 & 2.53 & 1.72 & 3.89 \\
        213 & 7.11 & 2.34 & 5.69 & 2.47 & 0.70 & 1.59 \\
        214 & 6.68 & 2.20 & 5.35 & 2.32 & 0.66 & 1.49 \\
        215 & 2.35 & 0.77 & 1.88 & 0.81 & 0.23 & 0.53 \\
        216 & 8.09 & 2.66 & 6.47 & 2.81 & 0.80 & 1.81 \\
        217 & 6.61 & 2.18 & 5.29 & 2.29 & 0.65 & 1.47 \\
        218 & 8.42 & 2.77 & 6.73 & 2.92 & 0.84 & 1.90 \\
        219 & 7.20 & 2.37 & 5.76 & 2.50 & 0.71 & 1.60 \\
        220 & 6.98 & 2.30 & 5.58 & 2.42 & 0.69 & 1.56 \\
        221 & 12.63 & 4.16 & 10.11 & 4.38 & 1.26 & 2.85 \\
        222 & 8.87 & 2.92 & 7.10 & 3.08 & 0.88 & 1.99 \\
        223 & 8.14 & 2.68 & 6.52 & 2.82 & 0.81 & 1.83 \\
        271 & 3.34 & 1.10 & 0.00 & 0.00 & 3.34 & 7.58 \\
        272 & 0.04 & 0.01 & 0.00 & 0.00 & 0.04 & 0.09 \\
        273 & 0.46 & 0.15 & 0.00 & 0.00 & 0.46 & 1.03 \\
        274 & 1.94 & 0.64 & 0.00 & 0.00 & 1.94 & 4.39 \\
        275 & 0.20 & 0.07 & 0.00 & 0.00 & 0.20 & 0.45 \\
        276 & 0.22 & 0.07 & 0.00 & 0.00 & 0.22 & 0.50 \\
        \hline
    \end{tabular}
\end{table}
\begin{table}
    \caption{Kind of post}
    \label{tab:data_per_post_kind}
    \centering
    \begin{tabular}{|l|S|S|S|S|S|S|}
        \hline
        Post Kind & {Total (h)} & {Total (\%)} & {Training (h)} & {Training(\%)} & {Test (h)} & {Test(\%)} \\
        \hline
        children's program & 0.4 & 0.1 & 0.4 & 0.2 & 0.0 & 0.0 \\
        church & 0.1 & 0.0 & 0.1 & 0.0 & 0.0 & 0.0 \\
        cinema/event tips & 1.0 & 0.3 & 0.7 & 0.3 & 0.2 & 0.4 \\
        comedy/sketch & 0.9 & 0.3 & 0.8 & 0.3 & 0.0 & 0.0 \\
        entertainment: other & 0.2 & 0.1 & 0.2 & 0.1 & 0.0 & 0.0 \\
        game moderation & 4.0 & 1.3 & 3.4 & 1.5 & 0.3 & 0.7 \\
        information & 120.5 & 39.7 & 88.5 & 38.4 & 20.9 & 47.3 \\
        live sports & 0.3 & 0.1 & 0.1 & 0.0 & 0.0 & 0.1 \\
        media info.: ext. & 0.5 & 0.2 & 0.3 & 0.1 & 0.1 & 0.3 \\
        moderation & 57.7 & 19.0 & 45.1 & 19.6 & 6.7 & 15.1 \\
        news & 74.4 & 24.5 & 57.3 & 24.8 & 10.2 & 23.1 \\
        parody/satire & 0.5 & 0.2 & 0.4 & 0.2 & 0.1 & 0.2 \\
        prog. info. (own) & 1.2 & 0.4 & 1.0 & 0.4 & 0.1 & 0.2 \\
        ref. to own digi. svc. & 1.7 & 0.6 & 1.4 & 0.6 & 0.2 & 0.5 \\
        service: other & 1.1 & 0.4 & 0.9 & 0.4 & 0.1 & 0.3 \\
        stock exchange & 0.5 & 0.2 & 0.3 & 0.1 & 0.0 & 0.1 \\
        traffic & 13.1 & 4.3 & 10.1 & 4.4 & 1.7 & 3.8 \\
        unknown & 4.6 & 1.5 & 3.5 & 1.5 & 0.7 & 1.5 \\
        weather & 20.8 & 6.9 & 16.1 & 7.0 & 2.8 & 6.3 \\
        \hline
    \end{tabular}
\end{table}
\begin{table}
    \caption{Journalistic form of the post}
    \label{tab:data_per_journalistic_form}
    \centering
    \begin{tabular}{|l|S|S|S|S|S|S|}
        \hline
        Journalistic Form & {Total (h)} & {Total (\%)} & {Training (h)} & {Training(\%)} & {Test (h)} & {Test(\%)} \\
        \hline
        (live) broadcast & 0.6 & 0.2 & 0.5 & 0.2 & 0.0 & 0.1 \\
        biography/ portrait & 1.5 & 0.5 & 1.3 & 0.6 & 0.1 & 0.2 \\
        commentary & 0.1 & 0.0 & 0.1 & 0.0 & 0.0 & 0.1 \\
        commentary/ column & 0.6 & 0.2 & 0.4 & 0.2 & 0.1 & 0.1 \\
        docu./feat./reportage & 19.3 & 6.4 & 14.8 & 6.4 & 2.4 & 5.5 \\
        explanatory film & 0.0 & 0.0 & 0.0 & 0.0 & 0.0 & 0.0 \\
        headline(s) & 2.9 & 1.0 & 2.2 & 1.0 & 0.4 & 1.0 \\
        interpret./ expl. piece & 0.2 & 0.1 & 0.1 & 0.0 & 0.1 & 0.1 \\
        message & 37.5 & 12.4 & 28.6 & 12.4 & 5.5 & 12.4 \\
        press rev. monothem. & 0.1 & 0.0 & 0.1 & 0.0 & 0.0 & 0.0 \\
        press rev.: var. topics & 0.2 & 0.1 & 0.2 & 0.1 & 0.0 & 0.0 \\
        report & 76.7 & 25.3 & 58.5 & 25.4 & 10.5 & 23.7 \\
        review/ criticism & 0.2 & 0.1 & 0.1 & 0.0 & 0.1 & 0.1 \\
        short report & 3.2 & 1.1 & 2.5 & 1.1 & 0.4 & 1.0 \\
        speaker's msg./ info & 7.5 & 2.5 & 6.0 & 2.6 & 0.8 & 1.9 \\
        studio convers./ talk & 40.7 & 13.4 & 30.7 & 13.3 & 6.7 & 15.1 \\
        unknown & 112.3 & 37.0 & 84.5 & 36.6 & 17.0 & 38.5 \\
        \hline
    \end{tabular}
\end{table}

\section{Personal Data and Offensive Content}
\label{app:pii}
The datasets may include personal information, such as the names of speakers or individuals related to the topic. Since the audio data is sourced from public broadcasts, we assume that the content is not offensive. Additionally, we have not received any complaints from the students who listened to the samples while reviewing the transcriptions.

\section{Potential Risks}
\label{app:risks}
The corpus was created to reflect diversity, aiming to include all dialect regions, while ensuring a balanced gender ratio. However, children and individuals over 65 might be significantly underrepresented, as most speakers are broadcast presenters, likely in their professional years. Thus, older individuals, as well as younger ones, are primarily represented in conversational contexts (ref. Table~\ref{tab:data_per_speaker}).
As a result, models trained on this dataset may perform below average for these demographic groups and individuals with strong, less commonly spoken dialects.

\section{Transcription Correction Rules}
\label{app:transcription_rules}
We established guidelines for correcting the automatically generated transcriptions to ensure consistency in the ground truth transcriptions. Additionally, we created a list where students could enter and search for the correct spelling of translated words and names (Table~\ref{tab:transcription_rules}).
\begin{table}
    \caption{Transcription rules for the manual corrections}
    \label{tab:transcription_rules}
    \centering
    \begin{tabular}{|l|p{8cm}|}
        \hline
         What& How\\
         \hline
         translations      & correct if clearly wrong and double-check with the list of corrected terms\\
         words             & correct if clearly wrong, missing, or too much\\
         names, dialect expressions & research correct words, and double-check the list of corrected terms\\
         end of sentence   & correct if necessary\\
         other punctuation & do not correct \\
         hyphens           & do not correct, do not add any when inserting text \\
         numbers           & leave as is, if the number (digits or words) itself is correct\\
         compounds         & do not correct as long as the meaning is correct \\
         tenses\tablefootnote{E.g Swiss German does not use past tense.}& do not correct \\
         capitalization    & correct if clearly wrong \\
         characters        & use Swiss characters only\tablefootnote{e.g. "ss" instead of "ß", which is used in Germany}\\
         repetitions, filler words\tablefootnote{Whisper generally does not transcribe fillers~\cite{ma_adapting_2023}} & leave as is, as long as it is also contained in the audio\\
         \hline
    \end{tabular}
\end{table}

\bibliographystyle{splncs04}
\bibliography{references}
\end{document}